\RequirePackage{snapshot}
\documentclass[12pt]{article}

\usepackage{amsmath}
\usepackage{amssymb}
\usepackage{amsfonts}
\usepackage{algorithm}
\usepackage{algorithmic}
\usepackage[font={small,sf},labelfont={small,sf,bf}]{caption}
\usepackage{graphicx}
\usepackage{color}
\usepackage{hyperref}
\usepackage[square,numbers]{natbib}
\bibpunct{(}{)}{;}{a}{,}{,}
\usepackage{cleveref}

\renewcommand\b{\begin{equation}}
\newcommand\e{\end{equation}}

\newcounter{mnote}

\newcommand{\flabel}[1]{\label{fig:#1}}
\newcommand{\seclabel}[1]{\label{sec:#1}}
\newcommand{\tlabel}[1]{\label{tab:#1}}
\newcommand{\elabel}[1]{\label{eq:#1}}

\newcommand{\fref}[1]{\Cref{fig:#1}}
\newcommand{\sref}[1]{\Cref{sec:#1}}
\newcommand{\tref}[1]{\Cref{tab:#1}}
\newcommand{\eref}[1]{\Cref{eq:#1}}

\newcommand*\idx[2][]
{ 
\def\next{#1}%
\ifx\empty\next
  (#2)
\else
  (#1, #2)
\fi
}
\newcommand*\elt[3][]
{ 
\def\next{#1}%
\ifx\empty\next
  #2\idx{#3}
\else
  #1\idx{#2,#3}
\fi
}
\newcommand*\pd[3][]
{ 
\def\next{#1}%
\ifx\empty\next
  \frac{\partial#2}{\partial #3}
\else
  \frac{{\partial^{#1} #2}}{\partial#3^{#1}}
\fi
}

\newcommand{\figdir}{Figures/}
\newcommand{\capt}[2]{\caption[#1.]{\textbf{#1.}#2}}

\newcommand{\fig}[5]
{
\begin{figure}
\begin{center}
\includegraphics[width=#3\columnwidth]{\figdir/#1}
\end{center}
\capt{#4}{#5}
\flabel{#2}
\end{figure}
}

\newcommand{\costwidth}{0.6375}
\newcommand{\spaceline}{\\[-2.2ex]\hline\\[-2.1ex]}

\author{
\begin{tabular}{l r}
Alex Graves & \hspace{1cm} \texttt{gravesa@google.com} \\
Greg Wayne  & \hspace{1cm}  \texttt{gregwayne@google.com} \\
Ivo Danihelka & \hspace{1cm}  \texttt{danihelka@google.com} \\ \\
\end{tabular}\\
Google DeepMind, London, UK
}
\date{}
\title{
\rule[0.4cm]{\textwidth}{2pt}
{\bf Neural Turing Machines}
\rule{\textwidth}{2pt} 
}

\begin{document}
\maketitle

\begin{center}
{\bf Abstract} 
\end{center}
We extend the capabilities of neural networks by coupling them to external memory resources, which they can interact with by attentional processes. The combined system is analogous to a Turing Machine or Von Neumann architecture but is differentiable end-to-end, allowing it to be efficiently trained with gradient descent. Preliminary results demonstrate that \textit{Neural Turing Machines} can infer simple algorithms such as copying, sorting, and associative recall from input and output examples. 

\section{Introduction}
Computer programs make use of three fundamental mechanisms: elementary operations (e.g., arithmetic operations), logical flow control (branching), and external memory, which can be written to and read from in the course of computation \citep{von1993first}. 
Despite its wide-ranging success in modelling complicated data, modern machine learning has largely neglected the use of logical flow control and external memory.

Recurrent neural networks (RNNs) stand out from other machine learning methods for their ability to learn and carry out complicated transformations of data over extended periods of time. Moreover, it is known that RNNs are Turing-Complete \citep{siegelmann1995turing}, and therefore have the capacity to simulate arbitrary procedures, \textit{if} properly wired. Yet what is possible in principle is not always what is simple in practice. We therefore enrich the capabilities of standard recurrent networks to simplify the solution of algorithmic tasks. This enrichment is primarily via a large, addressable memory, so, by analogy to Turing's enrichment of finite-state machines by an infinite memory tape, we dub our device a ``Neural Turing Machine'' (NTM). Unlike a Turing machine, an NTM is a differentiable computer that can be trained by gradient descent, yielding a practical mechanism for learning programs.

In human cognition, the process that shares the most similarity to algorithmic operation is known as ``working memory.'' While the mechanisms of working memory remain somewhat obscure at the level of neurophysiology, the verbal definition is understood to mean a capacity for short-term storage of information and its rule-based manipulation \citep{baddeley2009memory}. In computational terms, these rules are simple programs, and the stored information constitutes the arguments of these programs. Therefore, an NTM resembles a working memory system, as it is designed to solve tasks that require the application of approximate rules to ``rapidly-created variables.'' Rapidly-created variables \citep{hadley2009problem} are data that are quickly bound to memory slots, in the same way that the number \@3 and the number \@4 are put inside registers in a conventional computer and added to make \@7 \citep{minsky1967computation}. An NTM bears another close resemblance to models of working memory since the NTM architecture uses an attentional process to read from and write to memory selectively. In contrast to most models of working memory, our architecture can learn to use its working memory instead of deploying a fixed set of procedures over symbolic data. 

The organisation of this report begins with a brief review of germane research on working memory in psychology, linguistics, and neuroscience, along with related research in artificial intelligence and neural networks. We then describe our basic contribution, a memory architecture and attentional controller that we believe is well-suited to the performance of tasks that require the induction and execution of simple programs. To test this architecture, we have constructed a battery of problems, and we present their precise descriptions along with our results. We conclude by summarising the strengths of the architecture.

\section{Foundational Research}

\subsection{Psychology and Neuroscience}

The concept of working memory has been most heavily developed in psychology to explain the performance of tasks involving the short-term manipulation of information. The broad picture is that a ``central executive'' focuses attention and performs operations on data in a memory buffer \citep{baddeley2009memory}. Psychologists have extensively studied the capacity limitations of working memory, which is often quantified by the number of ``chunks'' of information that can be readily recalled \citep{miller1956magical}.\footnote{There remains vigorous debate about how best to characterise capacity limitations \citep{barrouillet2004time}.} These capacity limitations lead toward an understanding of structural constraints in the human working memory system, but in our own work we are happy to exceed them.

In neuroscience, the working memory process has been ascribed to the functioning of a system composed of the prefrontal cortex and basal ganglia \citep{goldman1995cellular}. Typical experiments involve recording from a single neuron or group of neurons in prefrontal cortex while a monkey is performing a task that involves observing a transient cue, waiting through a ``delay period,'' then responding in a manner dependent on the cue. Certain tasks elicit persistent firing from individual neurons during the delay period or more complicated neural dynamics. A recent study quantified delay period activity in prefrontal cortex for a complex, context-dependent task based on measures of ``dimensionality'' of the population code and showed that it predicted memory performance \citep{rigotti2013importance}.

Modeling studies of working memory range from those that consider how biophysical circuits could implement persistent neuronal firing \citep{wang1999synaptic} to those that try to solve explicit tasks \citep{hazy2006banishing} \citep{dayan2008simple} \citep{eliasmith2013build}. Of these, Hazy et al.'s model is the most relevant to our work, as it is itself analogous to the Long Short-Term Memory architecture, which we have modified ourselves. As in our architecture, Hazy et al.'s has mechanisms to gate information into memory slots, which they use to solve a memory task constructed of nested rules. In contrast to our work, the authors include no sophisticated notion of memory addressing, which limits the system to storage and recall of relatively simple, atomic data. Addressing, fundamental to our work, is usually left out from computational models in neuroscience, though it deserves to be mentioned that Gallistel and King \citep{gallistel2009memory} and Marcus \citep{marcus2003algebraic} have argued that addressing must be implicated in the operation of the brain.

\subsection{Cognitive Science and Linguistics}

Historically, cognitive science and linguistics emerged as fields at roughly the same time as artificial intelligence, all deeply influenced by the advent of the computer \citep{chomsky1956three} \citep{miller2003cognitive}. Their intentions were to explain human mental behaviour based on information or symbol-processing metaphors. In the early 1980s, both fields considered recursive or procedural (rule-based) symbol-processing to be the highest mark of cognition. The Parallel Distributed Processing (PDP) or connectionist revolution cast aside the symbol-processing metaphor in favour of a so-called ``sub-symbolic'' description of thought processes \citep{rumelhart1986parallel}.

Fodor and Pylyshyn \citep{fodor1988connectionism} famously made two barbed claims about the limitations of neural networks for cognitive modeling. They first objected that connectionist theories were incapable of \textit{variable-binding}, or the assignment of a particular datum to a particular slot in a data structure. In language, variable-binding is ubiquitous; for example, when one produces or interprets a sentence of the form, ``Mary spoke to John,'' one has assigned ``Mary'' the role of subject, ``John'' the role of object, and ``spoke to'' the role of the transitive verb. Fodor and Pylyshyn also argued that neural networks with fixed-length input domains could not reproduce human capabilities in tasks that involve processing \textit{variable-length structures}. In response to this criticism, neural network researchers including Hinton \citep{hinton1986learning}, Smolensky \citep{smolensky1990tensor}, Touretzky \citep{touretzky1990boltzcons}, Pollack \citep{pollack1990recursive}, Plate \citep{plate2003holographic}, and Kanerva \citep{kanerva2009hyperdimensional} investigated specific mechanisms that could support both variable-binding and variable-length structure within a connectionist framework. Our architecture draws on and potentiates this work.   

Recursive processing of variable-length structures continues to be regarded as a hallmark of human cognition. In the last decade, a firefight in the linguistics community staked several leaders of the field against one another. At issue was whether recursive processing is the ``uniquely human'' evolutionary innovation that enables language and is specialized to language, a view supported by Fitch, Hauser, and Chomsky \citep{fitch2005evolution}, or whether multiple new adaptations are responsible for human language evolution and recursive processing predates language \citep{jackendoff2005nature}. Regardless of recursive processing's evolutionary origins, all agreed that it is essential to human cognitive flexibility.

\subsection{Recurrent Neural Networks} \label{rnns}
Recurrent neural networks constitute a broad class of machines with dynamic state; that is, they have state whose evolution depends both on the input to the system and on the current state. In comparison to hidden Markov models, which also contain dynamic state, RNNs have a distributed state and therefore have significantly larger and richer memory and computational capacity. Dynamic state is crucial because it affords the possibility of context-dependent computation; a signal entering at a given moment can alter the behaviour of the network at a much later moment.

A crucial innovation to recurrent networks was the Long Short-Term Memory (LSTM) \citep{hochreiter1997long}. This very general architecture was developed for a specific purpose, to address the ``vanishing and exploding gradient'' problem \citep{hochreiter2001gradient}, which we might relabel the problem of ``vanishing and exploding sensitivity.'' LSTM ameliorates the problem by embedding perfect integrators \citep{sebastian1998continuous} for memory storage in the network. The simplest example of a perfect integrator is the equation $\mathbf{x}(t+1) = \mathbf{x}(t) + \mathbf{i}(t)$, where $\mathbf{i}(t)$ is an input to the system. The implicit identity matrix $I \mathbf{x}(t)$ means that signals do not dynamically vanish or explode. If we attach a mechanism to this integrator that allows an enclosing network to choose when the integrator listens to inputs, namely, a programmable gate depending on context, we have an equation of the form $\mathbf{x}(t+1) = \mathbf{x}(t) + g(\text{context}) \mathbf{i}(t)$. We can now selectively store information for an indefinite length of time. 

Recurrent networks readily process variable-length structures without modification. In sequential problems, inputs to the network arrive at different times, allowing variable-length or composite structures to be processed over multiple steps. Because they natively handle variable-length structures, they have recently been used in a variety of cognitive problems, including speech recognition \citep{graves2013speech,graves2014towards}, text generation \citep{sutskever2011generating}, handwriting generation \citep{graves2013generating} and machine translation \citep{sutskever2014sequence}. Considering this property, we do not feel that it is urgent or even necessarily valuable to build explicit parse trees to merge composite structures greedily \citep{pollack1990recursive} \citep{socher2012semantic} \citep{frasconi1998general}.

Other important precursors to our work include differentiable models of attention \citep{graves2013generating} \citep{BahdanauCB14} and program search \citep{hochreiter2001learning} \citep{das1992learning}, constructed with recurrent neural networks.

%
%
%

\section{Neural Turing Machines}
A Neural Turing Machine (NTM) architecture contains two basic components: a neural network \textit{controller} and a memory bank. \fref{ntm} presents a high-level diagram of the NTM architecture. Like most neural networks, the controller interacts with the external world via input and output vectors.
Unlike a standard network, it also interacts with a memory matrix using selective read and write operations.
By analogy to the Turing machine we refer to the network outputs that parametrise these operations as ``heads.'' 

\fig{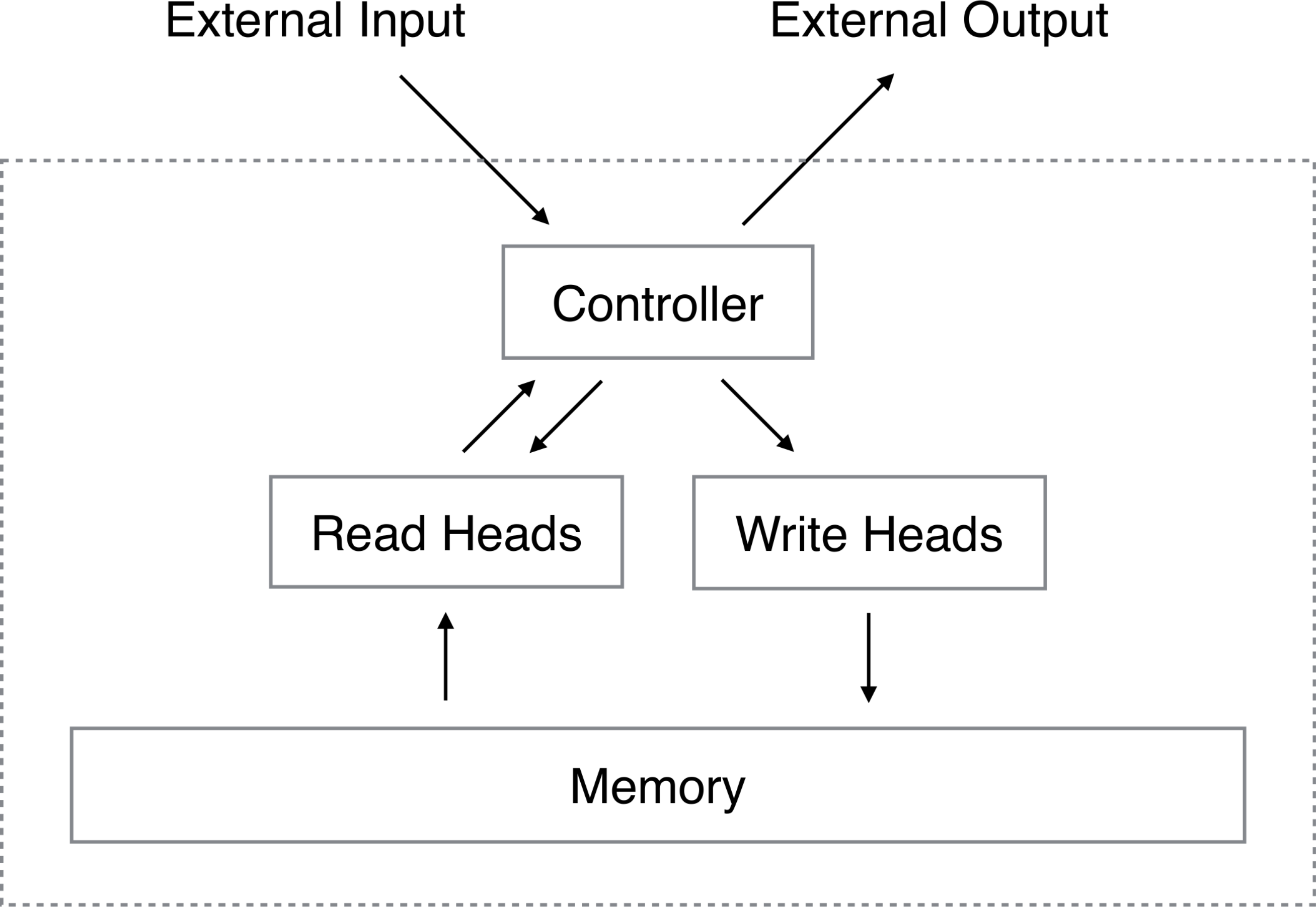}{ntm}{0.6}{Neural Turing Machine Architecture}{ During each update cycle, 
%
the controller network receives inputs from an external environment and emits outputs in response. It also reads to and writes from a memory matrix via a set of parallel read and write heads. The dashed line indicates the division between the NTM circuit and the outside world.}

Crucially, every component of the architecture is differentiable, making it straightforward to train with gradient descent. 
We achieved this by defining `blurry' read and write operations that interact to a greater or lesser degree with all the elements in memory (rather than addressing a single element, as in a normal Turing machine or digital computer).
The degree of blurriness is determined by an attentional ``focus'' mechanism that constrains each read and write operation to interact with a small portion of the memory, while ignoring the rest. Because interaction with the memory is highly sparse, the NTM is biased towards storing data without interference. The memory location brought into attentional focus is determined by specialised outputs emitted by the heads. These outputs define a normalised weighting over the rows in the memory matrix (referred to as memory ``locations''). Each weighting, one per read or write head, defines the degree to which the head reads or writes at each location. A head can thereby attend sharply to the memory at a single location or weakly to the memory at many locations.

\subsection{Reading}

Let $\mathbf{M}_t$ be the contents of the $N \times M$ memory matrix at time t, where $N$ is the number of memory locations, and $M$ is the vector size at each location.
Let $\mathbf{w}_t$ be a vector of weightings over the $N$ locations emitted by a read head at time $t$.
Since all weightings are normalised, the $N$ elements $w_t(i)$ of $\mathbf{w}_t$ obey the following constraints:
\begin{equation}
\sum_i w_t(i) = 1, \qquad 0 \leq w_t(i) \leq 1, \, \forall i.
\end{equation}
The length $M$ \textit{read vector} $\mathbf{r}_t$ returned by the head is defined as a convex combination of the row-vectors $\mathbf{M}_t(i)$ in memory:
\begin{equation}
\mathbf{r}_t \longleftarrow \sum_i{w_t(i) \mathbf{M}_t(i)},
\end{equation}
which is clearly differentiable with respect to both the memory and the weighting.

\subsection{Writing}
Taking inspiration from the input and forget gates in LSTM, we decompose each write into two parts: an \textit{erase} followed by an \textit{add}.


Given a weighting $\mathbf{w}_t$ emitted by a write head at time $t$, along with an \textit{erase vector} $\mathbf{e}_t$ whose $M$ elements all lie in the range $(0, 1)$, the memory vectors $\mathbf{M}_{t-1}(i)$ from the previous time-step are modified as follows:
\begin{equation}
\mathbf{\tilde{M}}_t(i) \longleftarrow \mathbf{M}_{t-1}(i) \left[\mathbf{1}-w_t(i) \mathbf{e}_t\right],    
\end{equation}
where $\mathbf{1}$ is a row-vector of all $1$-s, and the multiplication against the memory location acts point-wise.
Therefore, the elements of a memory location are reset to zero only if both the weighting at the location and the erase element are one; if either the weighting or the erase is zero, the memory is left unchanged. When multiple write heads are present, the erasures can be performed in any order, as multiplication is commutative.

Each write head also produces a length $M$ \textit{add vector} $\mathbf{a}_t$, which is added to the memory after the erase step has been performed:
\begin{equation}
\mathbf{M}_t(i) \longleftarrow \mathbf{\tilde{M}}_t(i) + w_t(i)\, \mathbf{a}_t. 
\end{equation}
Once again, the order in which the adds are performed by multiple heads is irrelevant.
The combined erase and add operations of all the write heads produces the final content of the memory at time $t$. Since both erase and add are differentiable, the composite write operation is differentiable too. Note that both the erase and add vectors have $M$ independent components, allowing fine-grained control over which elements in each memory location are modified.

\subsection{Addressing Mechanisms}
Although we have now shown the equations of reading and writing, we have not described how the weightings are produced. These weightings arise by combining two addressing mechanisms with complementary facilities. The first mechanism, ``content-based addressing,'' focuses attention on locations based on the similarity between their current values and values emitted by the controller. This is related to the content-addressing of Hopfield networks \citep{hopfield1982neural}. The advantage of content-based addressing is that retrieval is simple, merely requiring the controller to produce an approximation to a part of the stored data, which is then compared to memory to yield the exact stored value. 

However, not all problems are well-suited to content-based addressing. In certain tasks the content of a variable is arbitrary, but the variable still needs a recognisable name or address. Arithmetic problems fall into this category: the variable $x$ and the variable $y$ can take on any two values, but the procedure $f(x,y) = x \times y$ should still be defined. A controller for this task could take the values of the variables $x$ and $y$, store them in different addresses, then retrieve them and perform a multiplication algorithm. In this case, the variables are addressed by location, not by content. We call this form of addressing ``location-based addressing.'' Content-based addressing is strictly more general than location-based addressing as the content of a memory location could include location information inside it. In our experiments however, providing location-based addressing as a primitive operation proved essential for some forms of generalisation, so we employ both mechanisms concurrently.

\fref{address_flow_trim} presents a flow diagram of the entire addressing system that shows the order of operations for constructing a weighting vector when reading or writing.

\fig{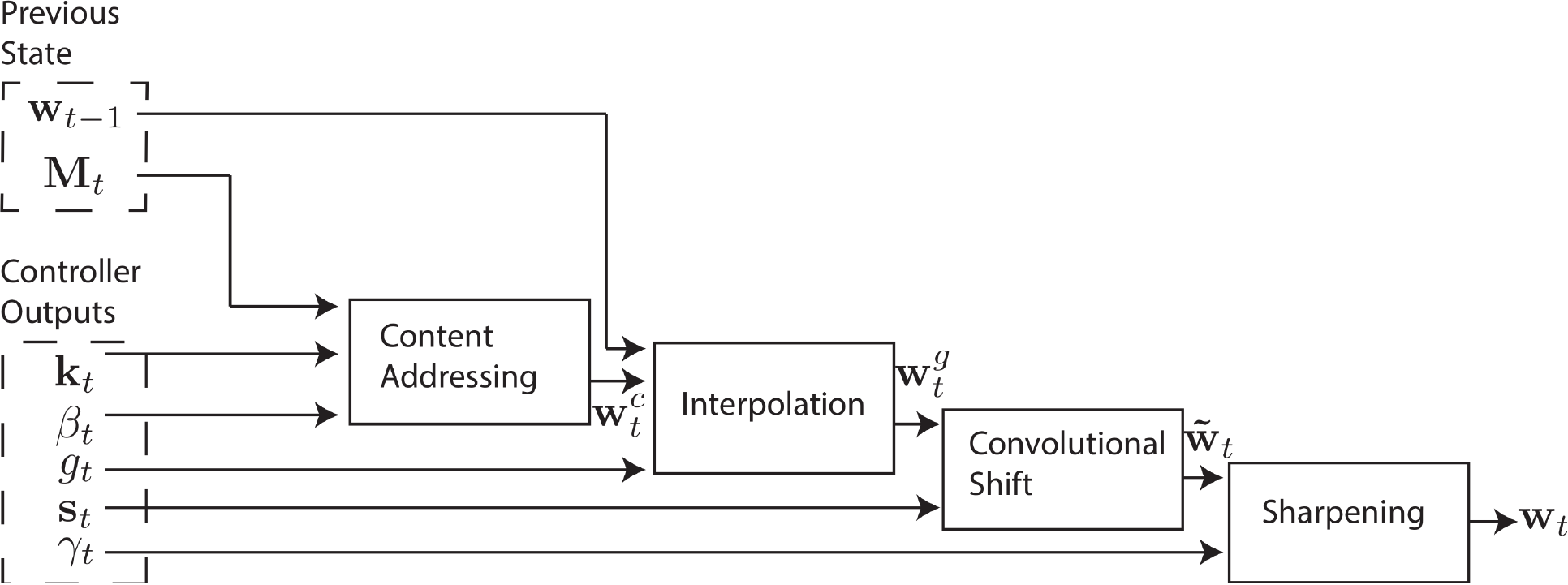}{address_flow_trim}{0.9}{Flow Diagram of the Addressing Mechanism}{ The \textit{key vector}, $\mathbf{k}_t$, and \textit{key strength}, $\beta_t$, are used to perform content-based addressing of the memory matrix, $\mathbf{M}_t$. The resulting content-based weighting is interpolated with the weighting from the previous time step based on the value of the \textit{interpolation gate}, $g_t$. The \textit{shift weighting}, $\mathbf{s}_t$, determines whether and by how much the weighting is rotated. Finally, depending on $\gamma_t$, the weighting is sharpened and used for memory access.}

\subsubsection{Focusing by Content}
For content-addressing, each head (whether employed for reading or writing) first produces a length $M$ \textit{key vector} $\mathbf{k}_t$ that is compared to each vector $\mathbf{M}_t(i)$ by a similarity measure $K\big[\cdot, \cdot\big]$. The content-based system produces a normalised weighting $w_t^c$ based on the similarity and a positive \textit{key strength}, $\beta_t$, which can amplify or attenuate the precision of the focus:
\begin{eqnarray} \label{content_focus}
w^c_t(i) & \longleftarrow &\frac{\exp\bigg(\beta_t K\big[\mathbf{k}_t, \mathbf{M}_t(i)\big]\bigg)}{\sum_j \exp\bigg(\beta_t K\big[\mathbf{k}_t, \mathbf{M}_t(j)\big]\bigg)}.
\end{eqnarray}
In our current implementation, the similarity measure is cosine similarity: 
\begin{eqnarray} \label{cosine}
K \big[ \mathbf{u}, \mathbf{v} \big] & = & \frac{\mathbf{u} \cdot \mathbf{v}}{||\mathbf{u}|| \cdot ||\mathbf{v}||}.
\end{eqnarray}

\subsubsection{Focusing by Location}\seclabel{location_addressing}
The location-based addressing mechanism is designed to facilitate both simple iteration across the locations of the memory and random-access jumps. 
It does so by implementing a rotational shift of a weighting. 
For example, if the current weighting focuses entirely on a single location, a rotation of $1$ would shift the focus to the next location. 
A negative shift would move the weighting in the opposite direction. 


%
%

Prior to rotation, each head emits a scalar \textit{interpolation gate} $g_t$ in the range $(0, 1)$.
The value of $g$ is used to blend between the weighting $\mathbf{w}_{t-1}$ produced by the head at the previous time-step and the weighting $\mathbf{w}^c_{t}$ produced by the content system at the current time-step, yielding the \textit{gated weighting} $\mathbf{w}^g_t$:
\begin{equation}\elabel{interpolation}
\mathbf{w}^g_t \longleftarrow g_t \mathbf{w}^c_{t}  + (1-g_t) \mathbf{w}_{t-1}.
\end{equation}
If the gate is zero, then the content weighting is entirely ignored, and the weighting from the previous time step is used.
Conversely, if the gate is one, the weighting from the previous iteration is ignored, and the system applies content-based addressing.


After interpolation, each head emits a \textit{shift weighting} $\mathbf{s}_t$ that defines a normalised distribution over the allowed integer shifts.
For example, if shifts between -1 and 1 are allowed, $\mathbf{s}_t$ has three elements corresponding to the degree to which shifts of -1, 0 and 1 are performed.
The simplest way to define the shift weightings is to use a softmax layer of the appropriate size attached to the controller.
We also experimented with another technique, where the controller emits a single scalar that is interpreted as the lower bound of a width one uniform distribution over shifts. For example, if the shift scalar is 6.7, then $s_t(6) = 0.3$, $s_t(7) = 0.7$, and the rest of $\mathbf{s}_t$ is zero.

If we index the $N$ memory locations from $0$ to $N-1$, the rotation applied to $\mathbf{w}^g_t$ by  $\mathbf{s}_t$ can be expressed as the following circular convolution:
\begin{align}\elabel{rotation}
\tilde{w}_t(i) 
&\longleftarrow \sum_{j=0}^{N-1} w^g_t(j)\, s_t(i-j)
\end{align}
where all index arithmetic is computed modulo $N$. 
The convolution operation in \eref{rotation} can cause leakage or dispersion of weightings over time if the shift weighting is not sharp.
For example, if shifts of -1, 0 and 1 are given weights of 0.1, 0.8 and 0.1, the rotation will transform a weighting focused at a single point into one slightly blurred over three points.
To combat this, each head emits one further scalar $\gamma_t \ge 1$ whose effect is to sharpen the final weighting as follows:
\begin{equation}
\elabel{sharpen}
w_t(i) \longleftarrow \frac{\tilde{w}_t(i)^{\gamma_t}}{\sum_j \tilde{w}_t(j)^{\gamma_t}}
\end{equation}

The combined addressing system of weighting interpolation and content and location-based addressing can operate in three complementary modes. 
One, a weighting can be chosen by the content system without any modification by the location system. Two, a weighting produced by the content addressing system can be chosen and then shifted. This allows the focus to jump to a location next to, but not on, an address accessed by content; in computational terms this allows a head to find a contiguous block of data, then access a particular element within that block. Three, a weighting from the previous time step can be rotated without any input from the content-based addressing system. 
This allows the weighting to iterate through a sequence of addresses by advancing the same distance at each time-step.

\subsection{Controller Network}\seclabel{controller}
The NTM architecture architecture described above has several free parameters, including the size of the memory, the number of read and write heads, and the range of allowed location shifts.
But perhaps the most significant architectural choice is the type of neural network used as the controller.
In particular, one has to decide whether to use a recurrent or feedforward network. 
A recurrent controller such as LSTM has its own internal memory that can complement the larger memory in the matrix. 
If one compares the controller to the central processing unit in a digital computer (albeit with adaptive rather than predefined instructions) and the memory matrix to RAM, then the hidden activations of the recurrent controller are akin to the registers in the processor.
They allow the controller to mix information across multiple time steps of operation. 
On the other hand a feedforward controller can mimic a recurrent network by reading and writing at the same location in memory at every step.
Furthermore, feedforward controllers often confer greater transparency to the network's operation because the pattern of reading from and writing to the memory matrix is usually easier to interpret than the internal state of an RNN. 
However, one limitation of a feedforward controller is that the number of concurrent read and write heads imposes a bottleneck on the type of computation the NTM can perform.
With a single read head, it can perform only a unary transform on a single memory vector at each time-step, with two read heads it can perform binary vector transforms, and so on. 
Recurrent controllers can internally store read vectors from previous time-steps, so do not suffer from this limitation.


\section{Experiments}
This section presents preliminary experiments on a set of simple algorithmic tasks such as copying and sorting data sequences.
The goal was not only to establish that NTM is able to solve the problems, but also that it is able to do so by learning compact internal programs.
The hallmark of such solutions is that they generalise well beyond the range of the training data.
For example, we were curious to see if a network that had been trained to copy sequences of length up to 20 could copy a sequence of length 100 with no further training.

For all the experiments we compared three architectures: NTM with a feedforward controller, NTM with an LSTM controller, and a standard LSTM network.
Because all the tasks were episodic, we reset the dynamic state of the networks at the start of each input sequence.
For the LSTM networks, this meant setting the previous hidden state equal to a learned bias vector.
For NTM the previous state of the controller, the value of the previous read vectors, and the contents of the memory were all reset to bias values.
All the tasks were supervised learning problems with binary targets; all networks had logistic sigmoid output layers and were trained with the cross-entropy objective function.
Sequence prediction errors are reported in bits-per-sequence.
For more details about the experimental parameters see \sref{details}.





\subsection{Copy}
The copy task tests whether NTM can store and recall a long sequence of arbitrary information. The network is presented with an input sequence of random binary vectors followed by a delimiter flag. Storage and access of information over long time periods has always been problematic for RNNs and other dynamic architectures. We were particularly interested to see if an NTM is able to bridge longer time delays than LSTM. 

The networks were trained to copy sequences of eight bit random vectors, where the sequence lengths were randomised between 1 and 20. 
The target sequence was simply a copy of the input sequence (without the delimiter flag).
Note that no inputs were presented to the network while it receives the targets, to ensure that it recalls the entire sequence with no intermediate assistance. 


%
As can be seen from \fref{copy_cost}, NTM (with either a feedforward or LSTM controller) learned much faster than LSTM alone, and converged to a lower cost. The disparity between the NTM and LSTM learning curves is dramatic enough to suggest a qualitative, rather than quantitative, difference in the way the two models solve the problem.

\fig{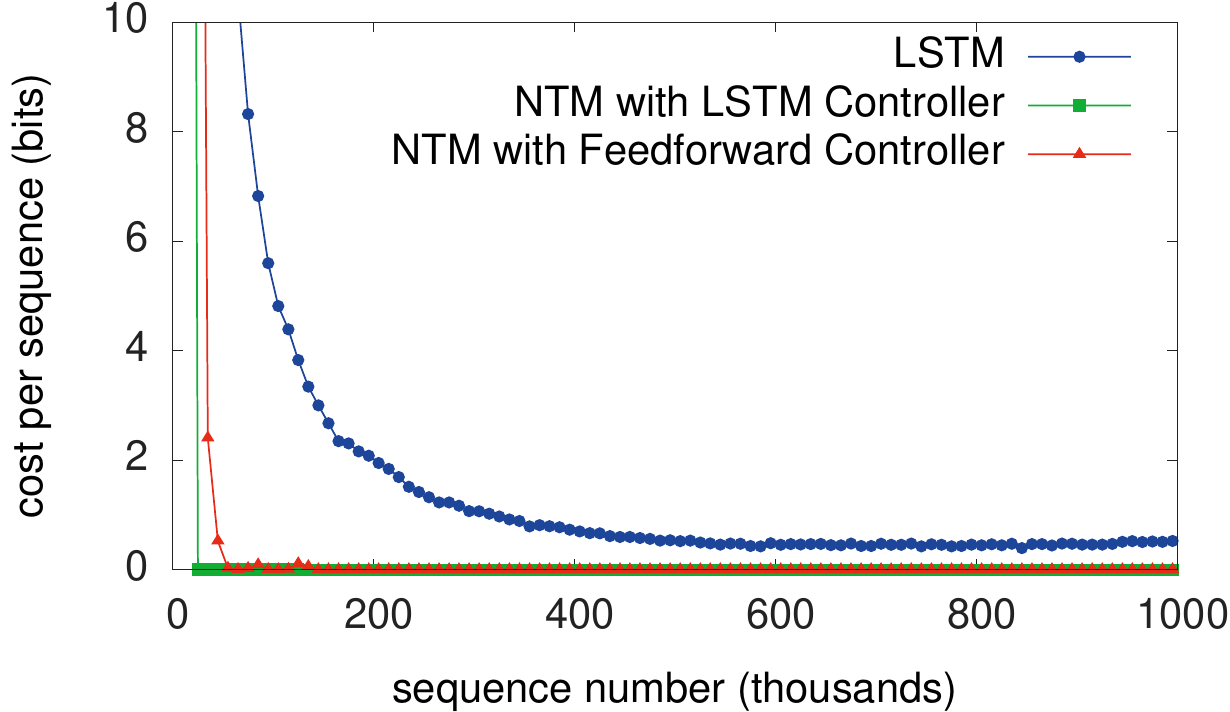}{copy_cost}{\costwidth}{Copy Learning Curves}{}

We also studied the ability of the networks to generalise to longer sequences than seen during training (that they can generalise to novel vectors is clear from the training error). 
\fref{copy_gen_ntm,fig:copy_gen_lstm} demonstrate that the behaviour of LSTM and NTM in this regime is radically different. NTM continues to copy as the length increases\footnote{The limiting factor was the size of the memory (128 locations), after which the cyclical shifts wrapped around and previous writes were overwritten.}, while LSTM rapidly degrades beyond length 20.
%
%
%



\fig{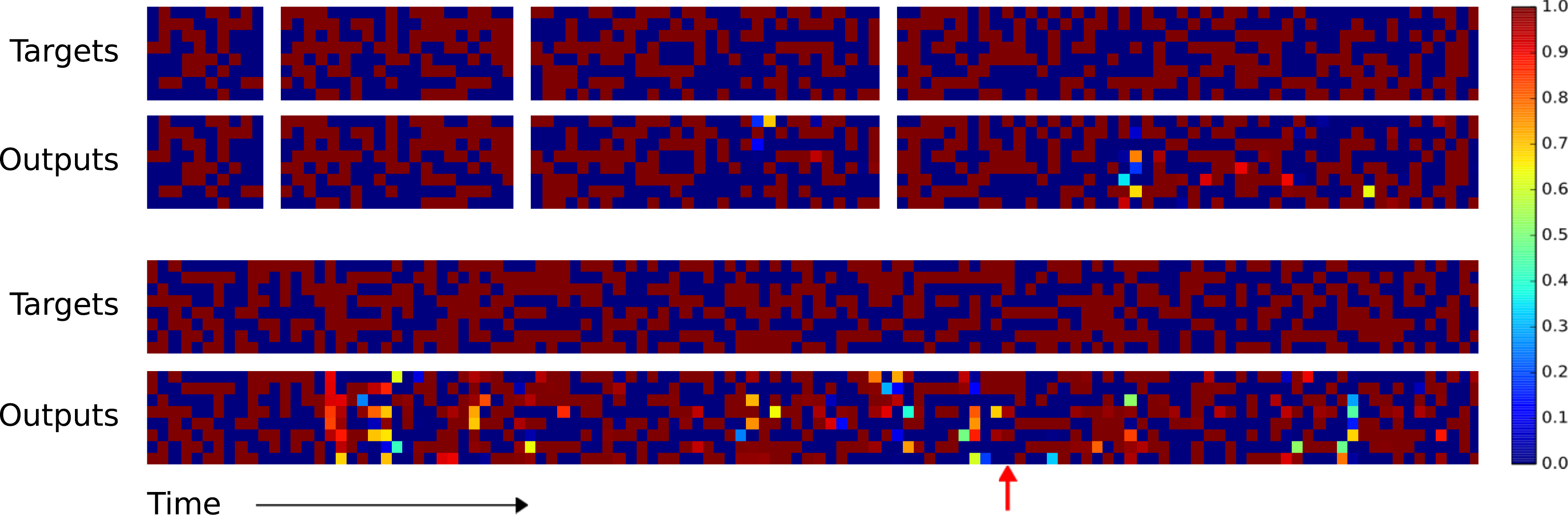}
{copy_gen_ntm}{0.9}{NTM Generalisation on the Copy Task}{ The four pairs of plots in the top row depict network outputs and corresponding copy targets for test sequences of length 10, 20, 30, and 50, respectively. The plots in the bottom row are for a length 120 sequence. The network was only trained on sequences of up to length 20. The first four sequences are reproduced with high confidence and very few mistakes. The longest one has a few more local errors and one global error: at the point indicated by the red arrow at the bottom, a single vector is duplicated, pushing all subsequent vectors one step back. Despite being subjectively close to a correct copy, this leads to a high loss.}


\fig{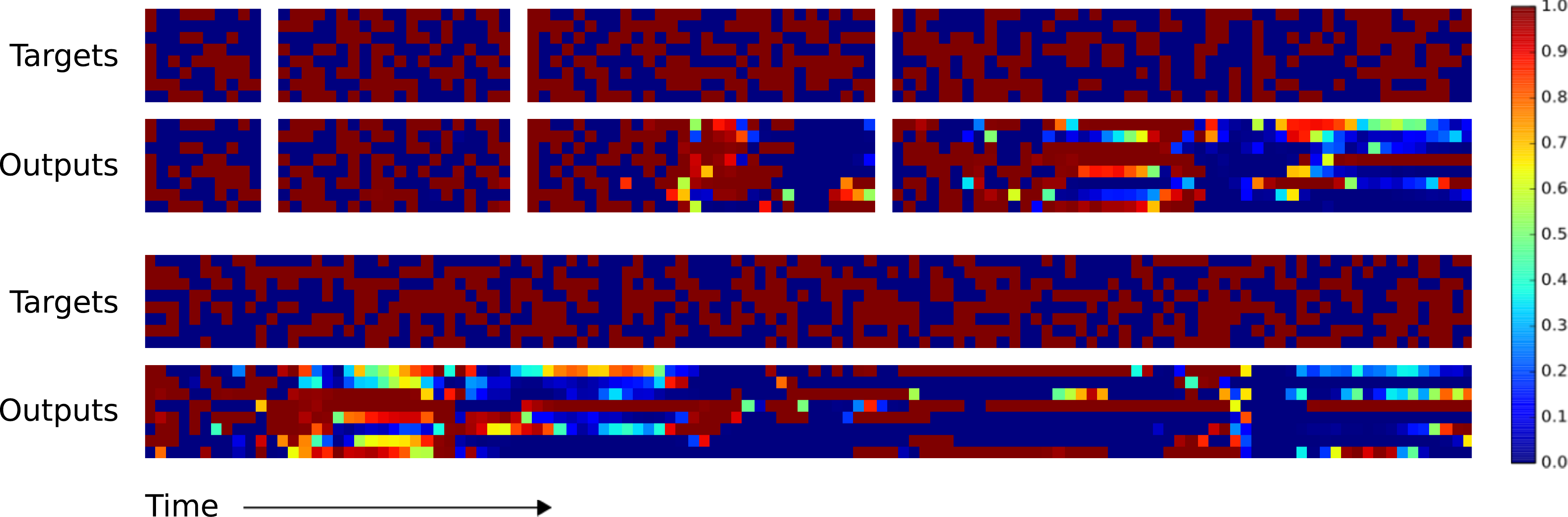}
{copy_gen_lstm}{0.9}{LSTM Generalisation on the Copy Task}{ The plots show inputs and outputs for the same sequence lengths as \fref{copy_gen_ntm}. Like NTM, LSTM learns to reproduce sequences of up to length 20 almost perfectly. However it clearly fails to generalise to longer sequences. Also note that the length of the accurate prefix decreases as the sequence length increases, suggesting that the network has trouble retaining information for long periods.}

The preceding analysis suggests that NTM, unlike LSTM, has learned some form of copy algorithm. 
To determine what this algorithm is, we examined the interaction between the controller and the memory (\fref{copy_mem}).
We believe that the sequence of operations performed by the network can be summarised by the following pseudocode:
\begin{algorithmic}
\STATE
\STATE \textbf{initialise:} move head to start location
\WHILE {input delimiter not seen}
\STATE receive input vector
\STATE write input to head location
\STATE increment head location by 1
\ENDWHILE
\STATE return head to start location
\WHILE {true}
\STATE read output vector from head location
\STATE emit output
\STATE increment head location by 1
\ENDWHILE
\STATE
\end{algorithmic}
\fig{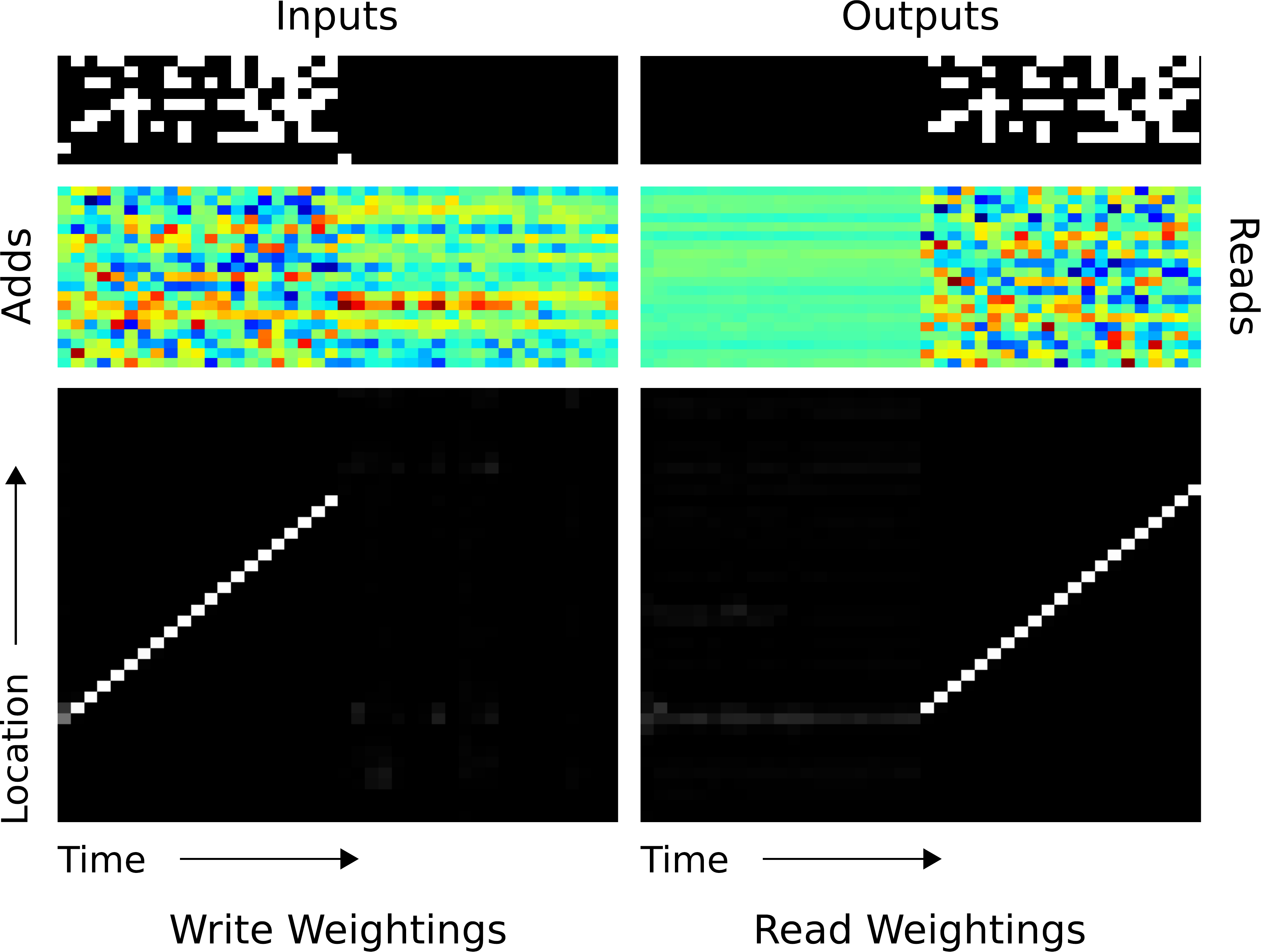}{copy_mem}{0.625}{NTM Memory Use During the Copy Task}{ 
The plots in the left column depict the inputs to the network (top), the vectors added to memory (middle) and the corresponding write weightings (bottom) during a single test sequence for the copy task. The plots on the right show the outputs from the network (top), the vectors read from memory (middle) and the read weightings (bottom).
Only a subset of memory locations are shown.
Notice the sharp focus of all the weightings on a single location in memory (black is weight zero, white is weight one).
Also note the translation of the focal point over time, reflects the network's use of iterative shifts for location-based addressing, as described in \sref{location_addressing}. Lastly, observe that the read locations exactly match the write locations, and the read vectors match the add vectors. This suggests that the network writes each input vector in turn to a specific memory location during the input phase, then reads from the same location sequence during the output phase.}



This is essentially how a human programmer would perform the same task in a low-level programming language. 
In terms of data structures, we could say that NTM has learned how to create and iterate through arrays.
Note that the algorithm combines both content-based addressing (to jump to start of the sequence) and location-based addressing (to move along the sequence).
Also note that the iteration would not generalise to long sequences without the ability to use relative shifts from the \textit{previous} read and write weightings (Equation~\ref{eq:interpolation}), and that without the focus-sharpening mechanism (Equation~\ref{eq:sharpen}) the weightings would probably lose precision over time.
%


\subsection{Repeat Copy}
The repeat copy task extends copy by requiring the network to output the copied sequence a specified number of times and then emit an end-of-sequence marker. 
The main motivation was to see if the NTM could learn a simple nested function. 
Ideally, we would like it to be able to execute a ``for loop'' containing any subroutine it has already learned.

The network receives random-length sequences of random binary vectors, followed by a scalar value indicating the desired number of copies, which appears on a separate input channel. 
To emit the end marker at the correct time the network must be both able to interpret the extra input and keep count of the number of copies it has performed so far. 
As with the copy task, no inputs are provided to the network after the initial sequence and repeat number. 
The networks were trained to reproduce sequences of size eight random binary vectors, where both the sequence length and the number of repetitions were chosen randomly from one to ten. 
The input representing the repeat number was normalised to have mean zero and variance one.
%

\fref{repeat_cost} shows that NTM learns the task much faster than LSTM, but both were able to solve it perfectly.\footnote{It surprised us that LSTM performed better here than on the copy problem. The likely reasons are that the sequences were shorter (up to length 10 instead of up to 20), and the LSTM network was larger and therefore had more memory capacity.} 
The difference between the two architectures only becomes clear when they are asked to generalise beyond the training data.
In this case we were interested in generalisation along two dimensions: sequence length and number of repetitions. 
\fref{repeat_gen} illustrates
the effect of doubling first one, then the other, for both LSTM and NTM.
Whereas LSTM fails both tests, NTM succeeds with longer sequences and is able to perform more than ten repetitions;
however it is unable to keep count of of how many repeats it has completed, and does not predict the end marker correctly.
This is probably a consequence of representing the number of repetitions numerically, which does not easily generalise beyond a fixed range.

\fref{repeat_mem} suggests that NTM learns a simple extension of the copy algorithm in the previous section, where the sequential read is repeated as many times as necessary.
%
%
%
%
%
%
%
\fig{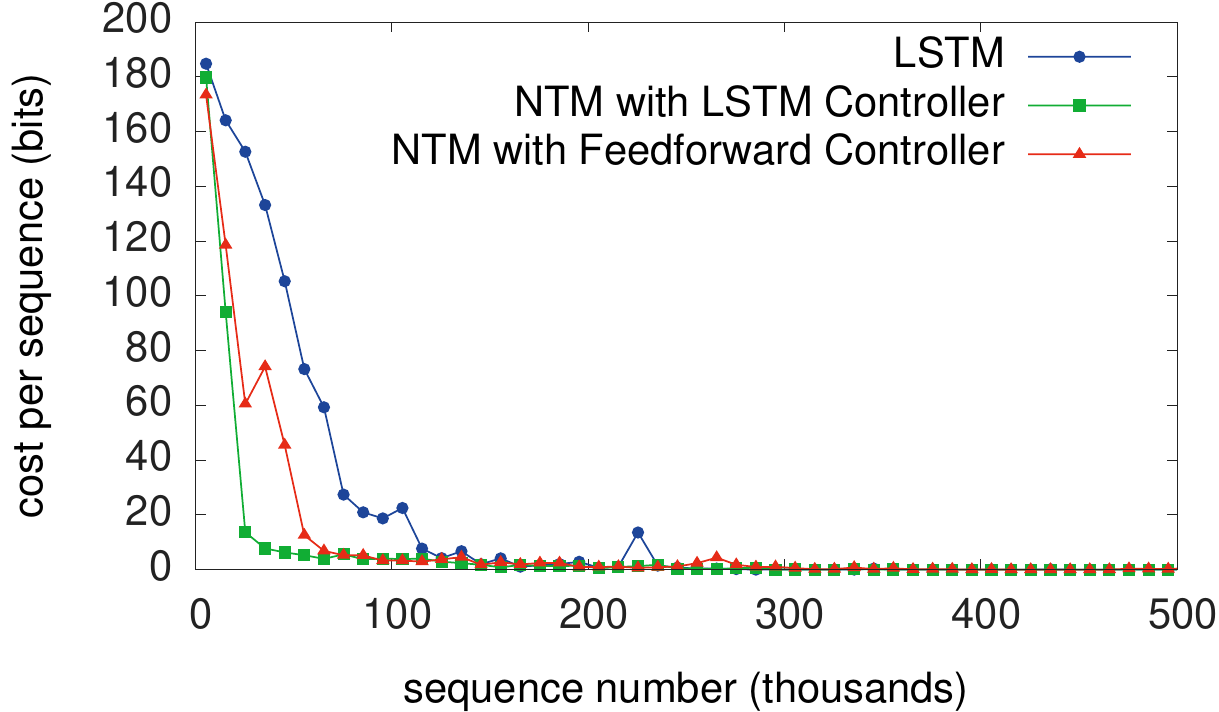}{repeat_cost}{\costwidth}{Repeat Copy Learning Curves}{}
%
%
%
\fig{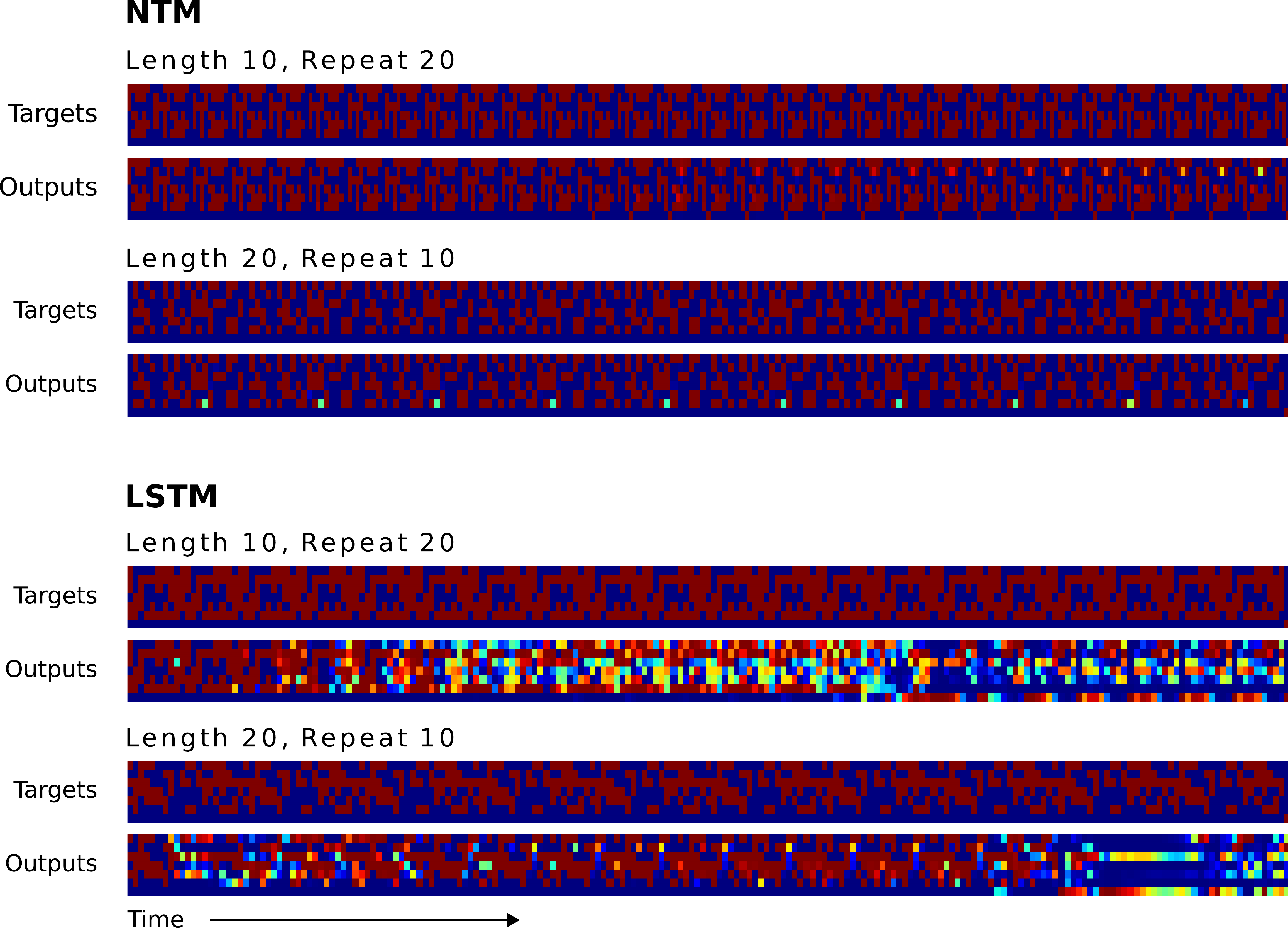}
{repeat_gen}{0.9}{NTM and LSTM Generalisation for the Repeat Copy Task}{ NTM generalises almost perfectly to longer sequences than seen during training. When the number of repeats is increased it is able to continue duplicating the input sequence fairly accurately; but it is unable to predict when the sequence will end, emitting the end marker after the end of every repetition beyond the eleventh. LSTM struggles with both increased length and number, rapidly diverging from the input sequence in both cases.}

\fig{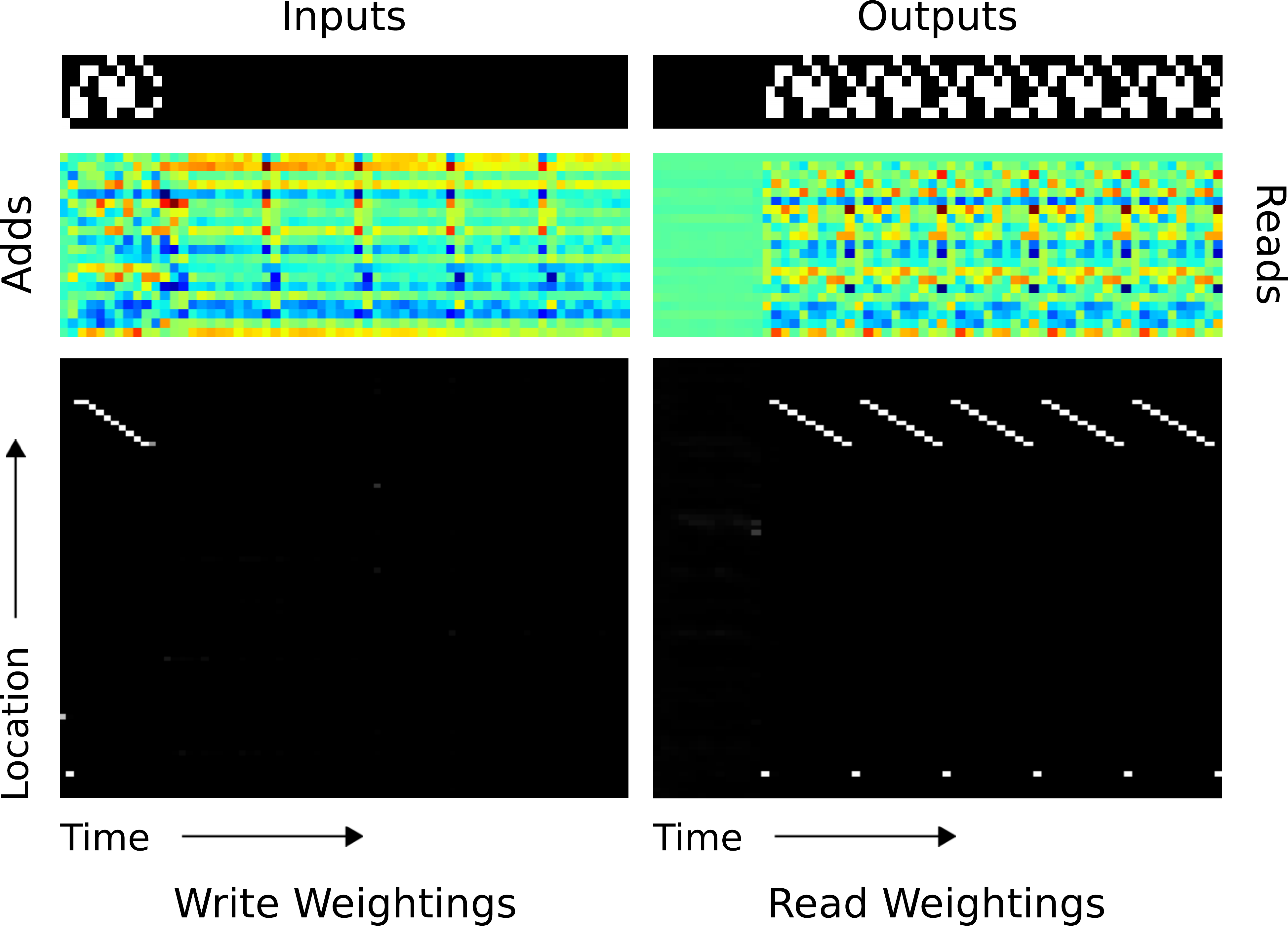}{repeat_mem}{0.7}{NTM Memory Use During the Repeat Copy Task}{ 
As with the copy task the network first writes the input vectors to memory using iterative shifts. It then reads through the sequence to replicate the input as many times as necessary (six in this case). The white dot at the bottom of the read weightings seems to correspond to an intermediate location used to redirect the head to the start of the sequence (The NTM equivalent of a \textit{goto} statement).}


\subsection{Associative Recall}

The previous tasks show that the NTM can apply algorithms to relatively simple, linear data structures. The next order of complexity in organising data arises from ``indirection''---that is, when one data item points to another. We test the NTM's capability for learning an instance of this more interesting class by constructing a list of items so that querying with one of the items demands that the network return the subsequent item. More specifically, we define an item as a sequence of binary vectors that is bounded on the left and right by delimiter symbols. After several items have been propagated to the network, we query by showing a random item, and we ask the network to produce the next item. In our experiments, each item consisted of three six-bit binary vectors (giving a total of 18 bits per item). During training, we used a minimum of 2 items and a maximum of 6 items in a single episode. 

\fref{assoc_cost} shows that NTM learns this task significantly faster than LSTM, terminating at near zero cost within approximately $30,000$ episodes, whereas LSTM does not reach zero cost after a million episodes. Additionally, NTM with a feedforward controller learns faster than NTM with an LSTM controller. 
These two results suggest that NTM's external memory is a more effective way of maintaining the data structure than LSTM's internal state.
NTM also generalises much better to longer sequences than LSTM, as can be seen in \fref{assoc_gen}. NTM with a feedforward controller is nearly perfect for sequences of up to 12 items (twice the maximum length used in training), and still has an average cost below 1 bit per sequence for sequences of 15 items.


\fig{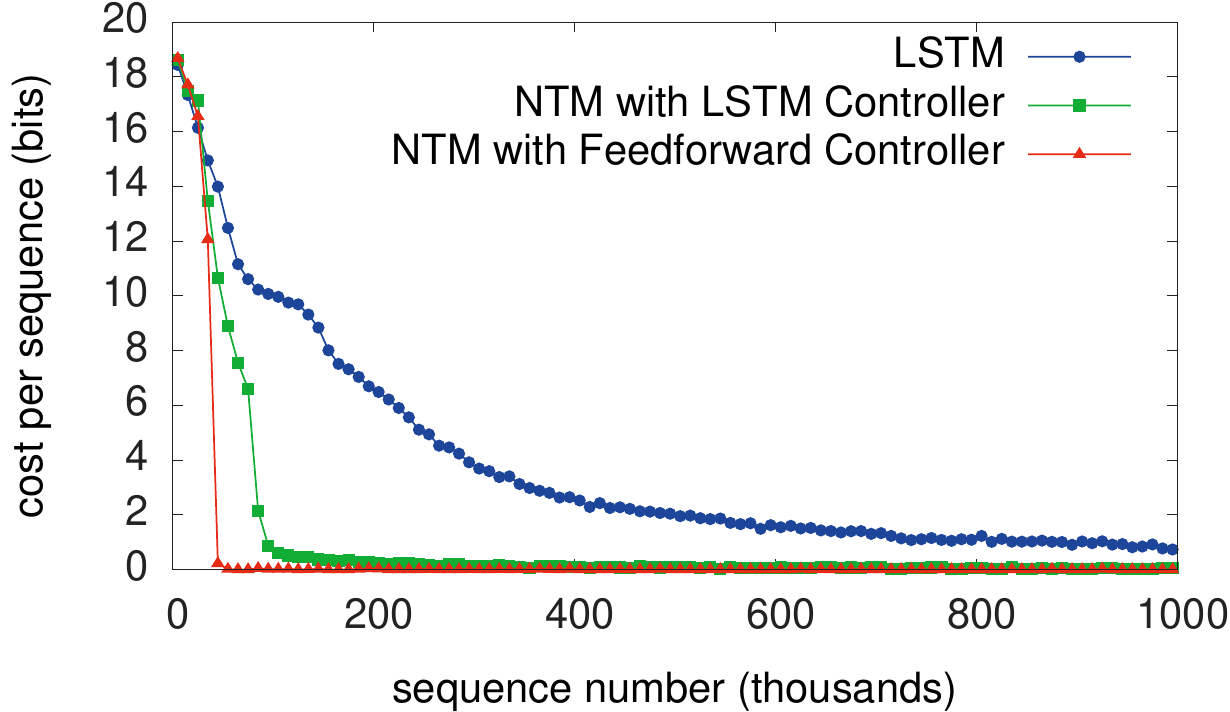}{assoc_cost}{\costwidth}{Associative Recall Learning Curves for NTM and LSTM}{}
\fig{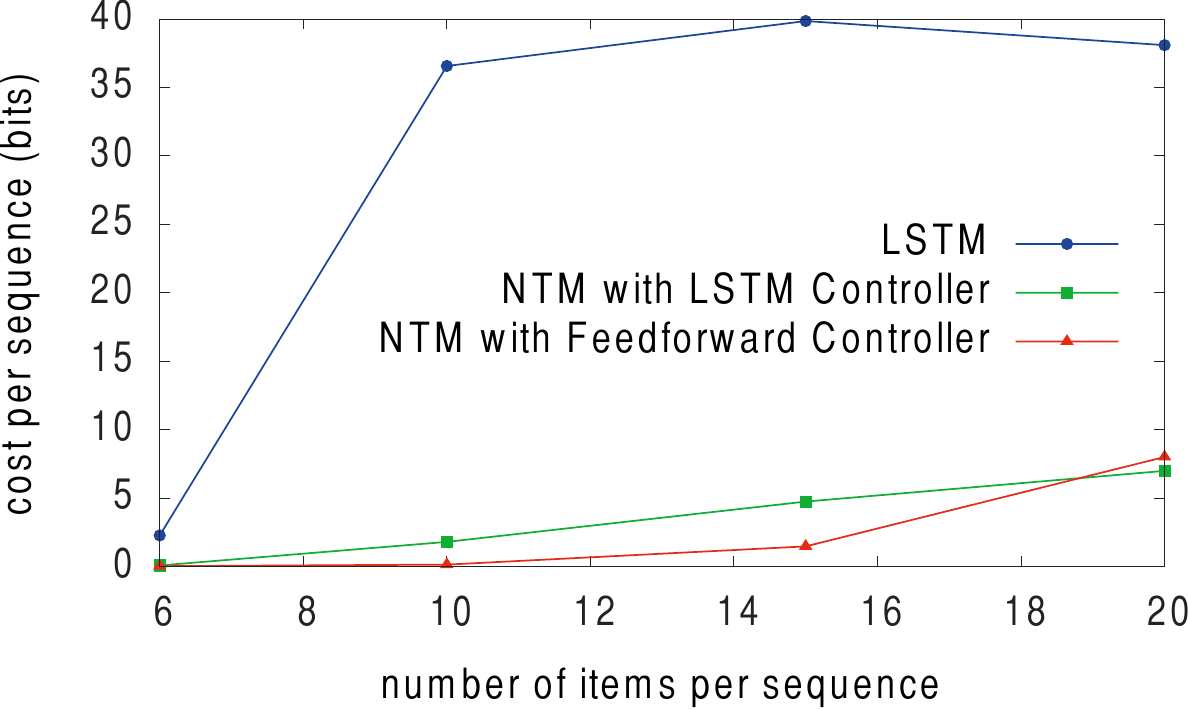}{assoc_gen}{0.6}{Generalisation Performance on Associative Recall for Longer Item Sequences}{ The NTM with either a feedforward or LSTM controller generalises to much longer sequences of items than the LSTM alone. In particular, the NTM with a feedforward controller is nearly perfect for item sequences of twice the length of sequences in its training set.}

In \fref{assoc_mem}, we show the operation of the NTM memory, controlled by an LSTM with one head, on a single test episode. In ``Inputs,'' we see that the input denotes item delimiters as single bits in row 7. After the sequence of items has been propagated, a delimiter in row 8 prepares the network to receive a query item. In this case, the query item corresponds to the second item in the sequence (contained in the green box). In ``Outputs,'' we see that the network crisply outputs item 3 in the sequence (from the red box). In ``Read Weightings,'' on the last three time steps, we see that the controller reads from contiguous locations that each store the time slices of item 3. This is curious because it appears that the network has jumped directly to the correct location storing item 3. However we can explain this behaviour by looking at ``Write Weightings.'' Here we see that the memory is written to even when the input presents a delimiter symbol between items. One can confirm in ``Adds'' that data are indeed written to memory when the delimiters are presented (e.g., the data within the black box); furthermore, each time a delimiter is presented, the vector added to memory is different. Further analysis of the memory reveals that the network accesses the location it reads after the query by using a content-based lookup that produces a weighting that is shifted by one. Additionally, the key used for content-lookup corresponds to the vector that was added in the black box. This implies the following memory-access algorithm: when each item delimiter is presented, the controller writes a compressed representation of the previous three time slices of the item. After the query arrives, the controller recomputes the same compressed representation of the query item, uses a content-based lookup to find the location where it wrote the first representation, and then shifts by one to produce the subsequent item in the sequence (thereby combining content-based lookup with location-based offsetting).

\fig{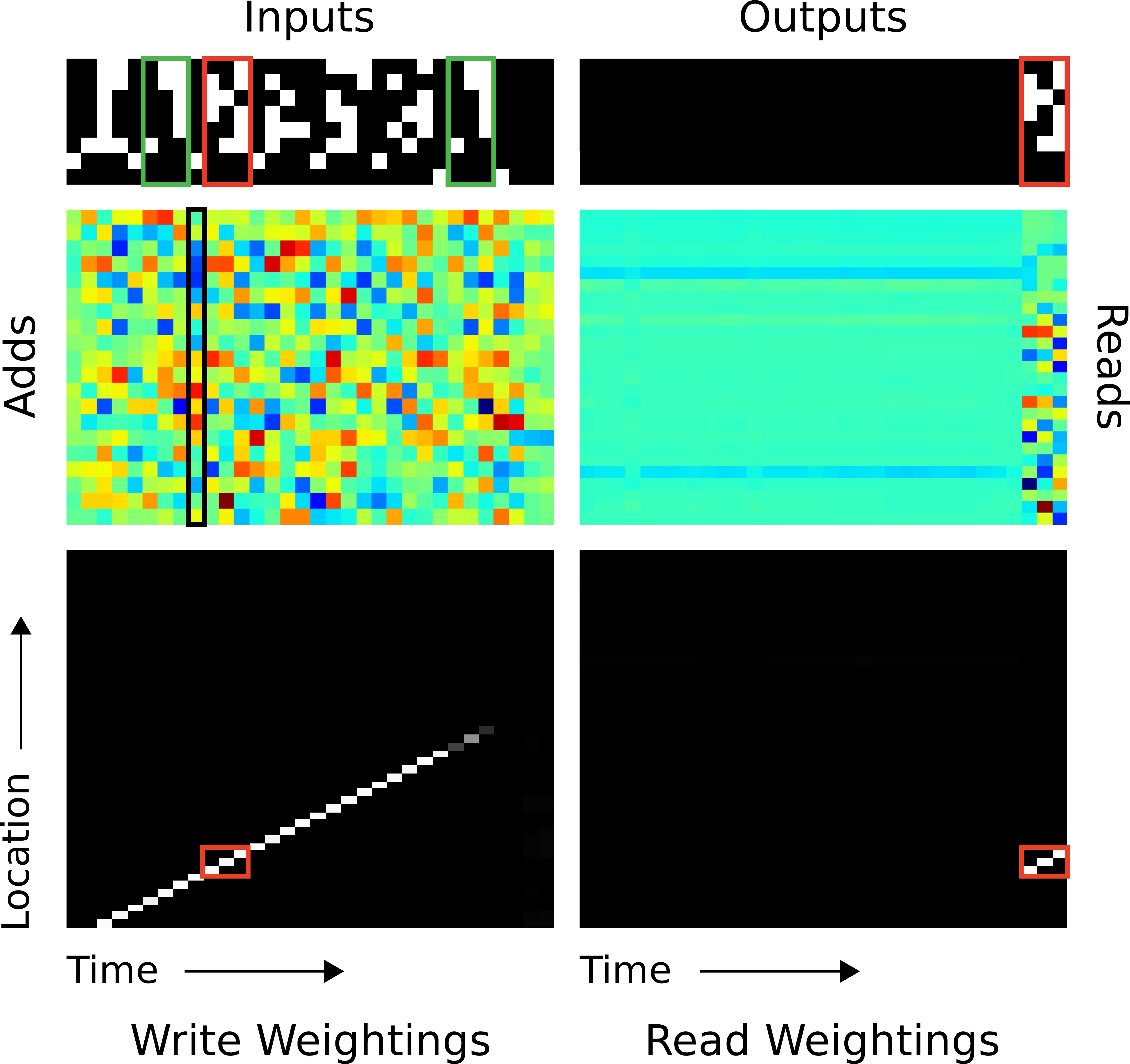}{assoc_mem}{0.6}{NTM Memory Use During the Associative Recall Task}{ In ``Inputs,'' a sequence of items, each composed of three consecutive binary random vectors is propagated to the controller. The distinction between items is designated by delimiter symbols (row 7 in ``Inputs''). After several items have been presented, a delimiter that designates a query is presented (row 8 in ``Inputs''). A single query item is presented (green box), and the network target corresponds to the subsequent item in the sequence (red box). 
In ``Outputs,'' we see that the network correctly produces the target item. 
The red boxes in the read and write weightings highlight the three locations where the target item was written and then read.
The solution the network finds is to form a compressed representation (black box in ``Adds'') of each item that it can store in a single location. For further analysis, see the main text.}


\subsection{Dynamic N-Grams}
The goal of the dynamic N-Grams task was to test whether NTM could rapidly adapt to new predictive distributions.
In particular we were interested to see if it were able to use its memory as a re-writable table that it could use to keep count of transition statistics, thereby emulating a conventional N-Gram model.

We considered the set of all possible 6-Gram distributions over binary sequences.
Each 6-Gram distribution can be expressed as a table of $2^5=32$ numbers, specifying the probability that the next bit will be one, given all possible length five binary histories.
For each training example, we first generated random 6-Gram probabilities by independently drawing all $32$ probabilities from the $Beta(\frac{1}{2}, \frac{1}{2})$ distribution.

We then generated a particular training sequence by drawing 200 successive bits using the current lookup table.\footnote{The first 5 bits, for which insufficient context exists to sample from the table, are drawn i.i.d. from a Bernoulli distribution with $p=0.5$.}
The network observes the sequence one bit at a time and is then asked to predict the next bit.
The optimal estimator for the problem can be determined by Bayesian analysis \citep{murphy2012machine}:
\begin{equation}
P(B=1|N_1, N_0, \mathbf{c}) = \frac{N_1 + \frac{1}{2}}{N_1 + N_0 + 1}
\end{equation}
where $\mathbf{c}$ is the five bit previous context, $B$ is the value of the next bit and $N_0$ and $N_1$ are respectively the number of zeros and ones observed after $\mathbf{c}$ so far in the sequence.
We can therefore compare NTM to the optimal predictor as well as LSTM.
To assess performance we used a validation set of 1000 length 200 sequences sampled from the same distribution as the training data.
As shown in \fref{ngram_cost}, NTM achieves a small, but significant performance advantage over LSTM, but never quite reaches the optimum cost.

\fig{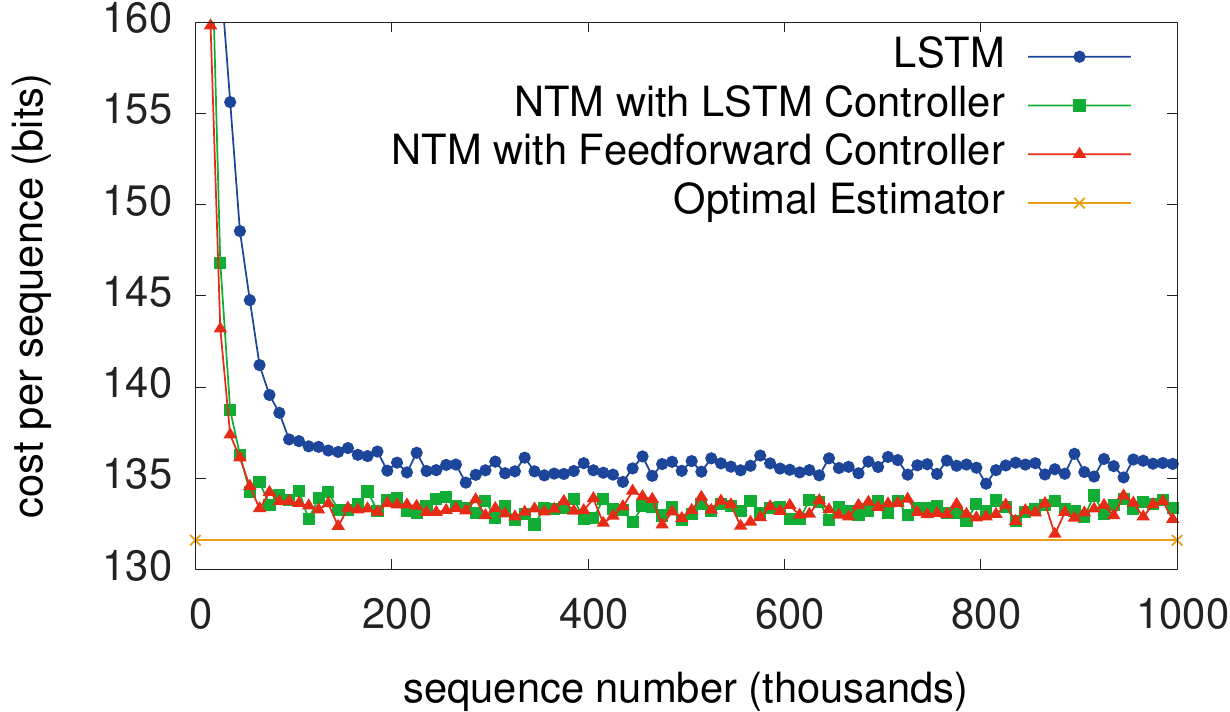}{ngram_cost}{\costwidth}{Dynamic N-Gram Learning Curves}{}

The evolution of the two architecture's predictions as they observe new inputs is shown in \fref{ngram_inference}, along with the optimal predictions.
Close analysis of NTM's memory usage (\fref{ngram_mem}) suggests that the controller uses the memory to count how many ones and zeros it has observed in different contexts, allowing it to implement an algorithm similar to the optimal estimator.


\fig{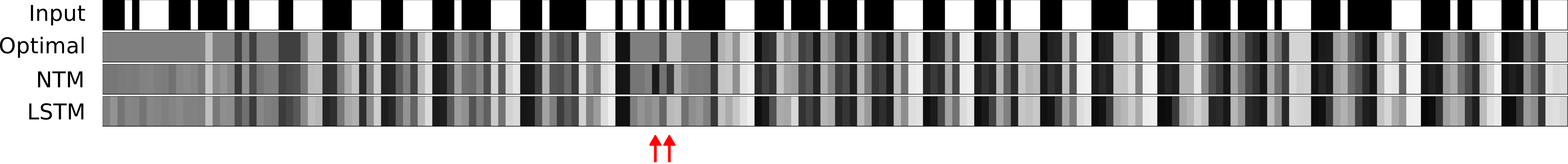}{ngram_inference}{1}{Dynamic N-Gram Inference}{ The top row shows a test sequence from the N-Gram task, and the rows below show the corresponding predictive distributions emitted by the optimal estimator, NTM, and LSTM. In most places the NTM predictions are almost indistinguishable from the optimal ones. However at the points indicated by the two arrows it makes clear mistakes, one of which is explained in \fref{ngram_mem}. LSTM follows the optimal predictions closely in some places but appears to diverge further as the sequence progresses; we speculate that this is due to LSTM ``forgetting'' the observations at the start of the sequence.}

\fig{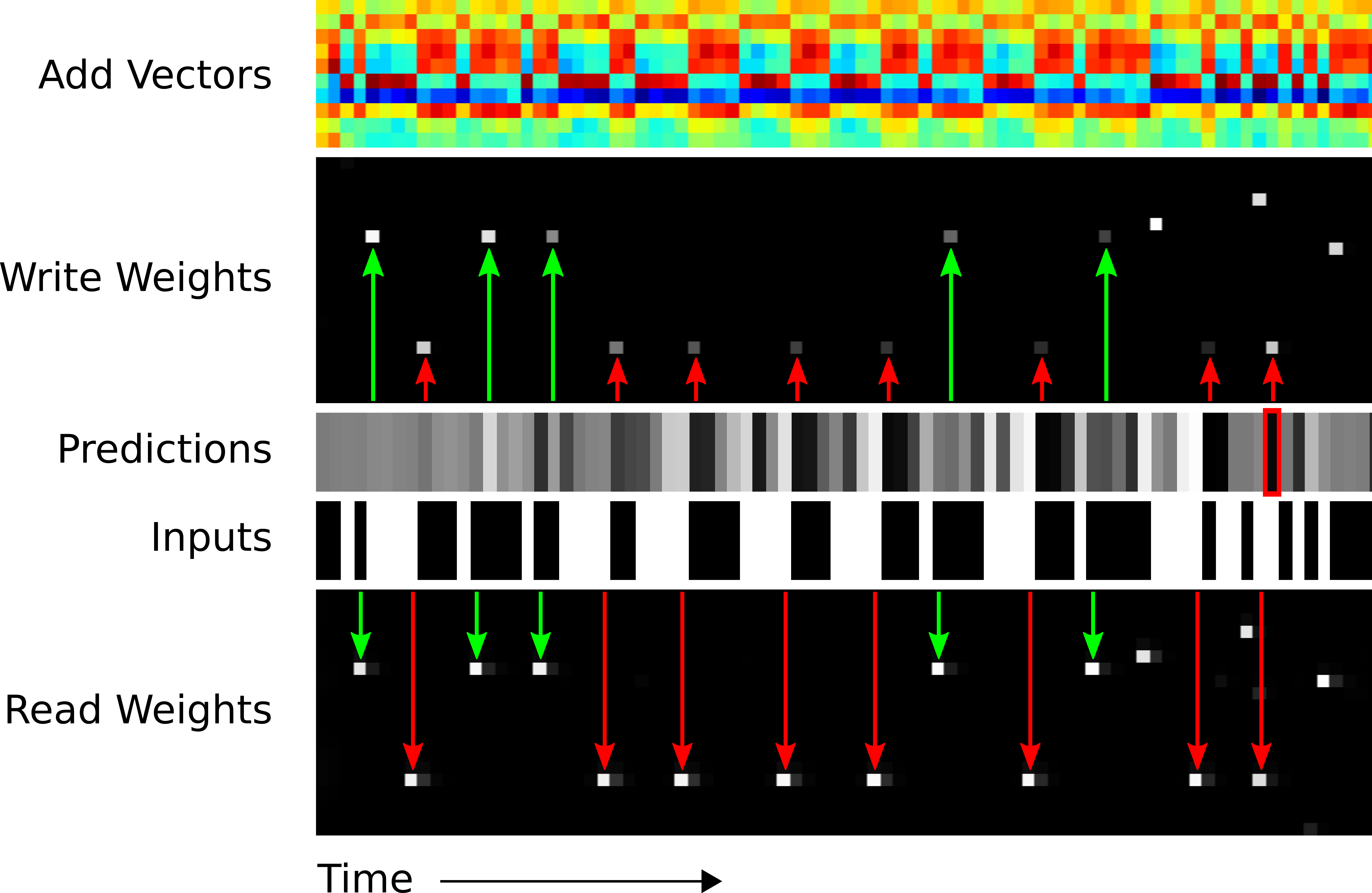}{ngram_mem}{0.725}{NTM Memory Use During the Dynamic N-Gram Task}{ The red and green arrows indicate point where the same context is repeatedly observed during the test sequence (``00010'' for the green arrows, ``01111'' for the red arrows). At each such point the same location is accessed by the read head, and then, on the next time-step, accessed by the write head. We postulate that the network uses the writes to keep count of the fraction of ones and zeros following each context in the sequence so far. This is supported by the add vectors, which are clearly anti-correlated at places where the input is one or zero, suggesting a distributed ``counter.'' Note that the write weightings grow fainter as the same context is repeatedly seen; this may be because the memory records a ratio of ones to zeros, rather than absolute counts. The red box in the prediction sequence corresponds to the mistake at the first red arrow in \fref{ngram_inference}; the controller appears to have accessed the wrong memory location, as the previous context was ``01101'' and not ``01111.''}

\subsection{Priority Sort}
This task tests whether the NTM can sort data---an important elementary algorithm. A sequence of random binary vectors is input to the network along with a scalar priority rating for each vector. The priority is drawn uniformly from the range [-1, 1]. The target sequence contains the binary vectors sorted according to their priorities, as depicted in \fref{sort_outputs}.

\fig{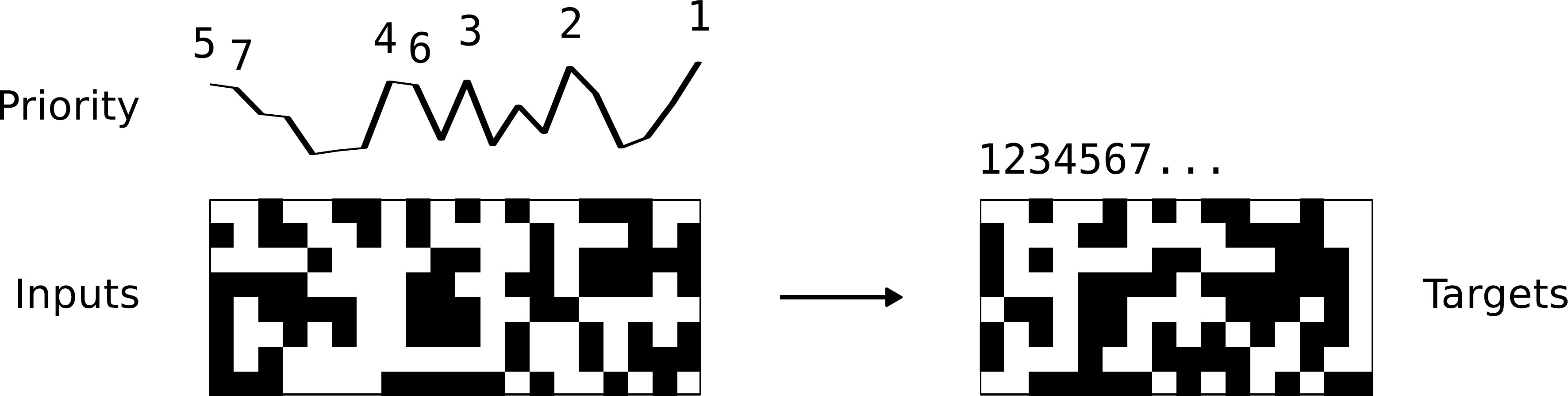}{sort_outputs}{0.75}{Example Input and Target Sequence for the Priority Sort Task}{ The input sequence contains random binary vectors and random scalar priorities. The target sequence is a subset of the input vectors sorted by the priorities.}

Each input sequence contained 20 binary vectors with corresponding priorities, and each target sequence was the 16 highest-priority vectors in the input.\footnote{We limited the sort to size 16 because we were interested to see if NTM would solve the task using a binary heap sort of depth 4.} 
Inspection of NTM's memory use led us to hypothesise that it uses the priorities to determine the relative location of each write.
To test this hypothesis we fitted a linear function of the priority to the observed write locations.
\fref{sort_mem} shows that the locations returned by the linear function closely match the observed write locations. It also shows that the network reads from the memory locations in increasing order, thereby traversing the sorted sequence.

The learning curves in \fref{sort_cost} demonstrate that NTM with both feedforward and LSTM controllers substantially outperform LSTM on this task.
Note that eight parallel read and write heads were needed for best performance with a feedforward controller on this task; this may reflect the difficulty of sorting vectors using only unary vector operations (see \sref{controller}).

\fig{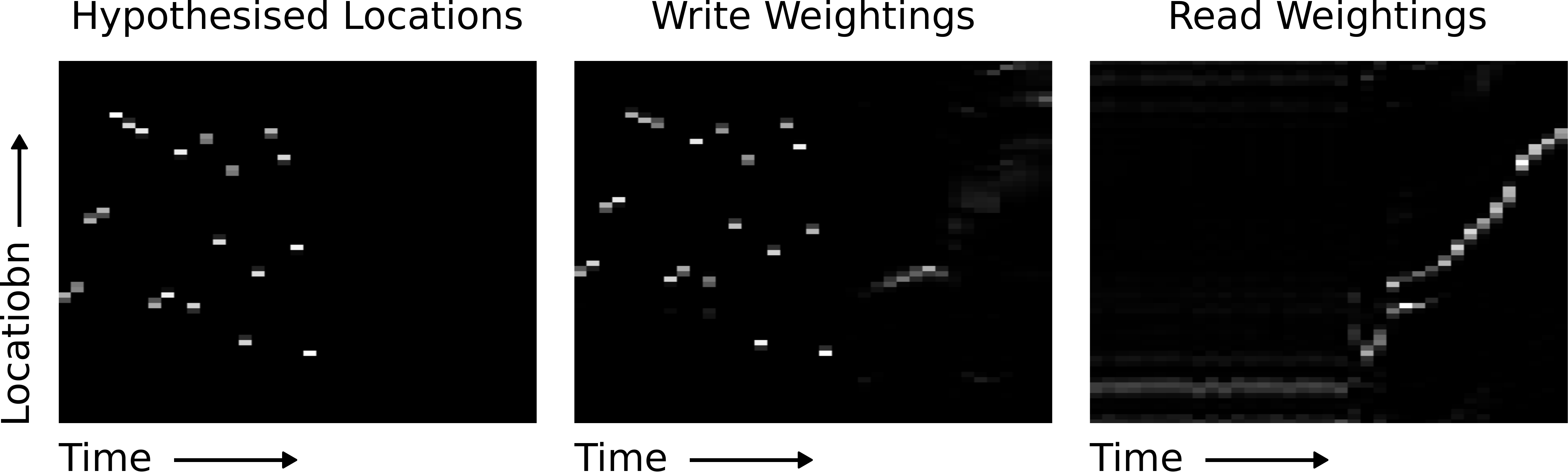}{sort_mem}{0.9}{NTM Memory Use During the Priority Sort Task}{ Left: Write locations returned by fitting a linear function of the priorities to the observed write locations. Middle: Observed write locations. Right: Read locations.}

\fig{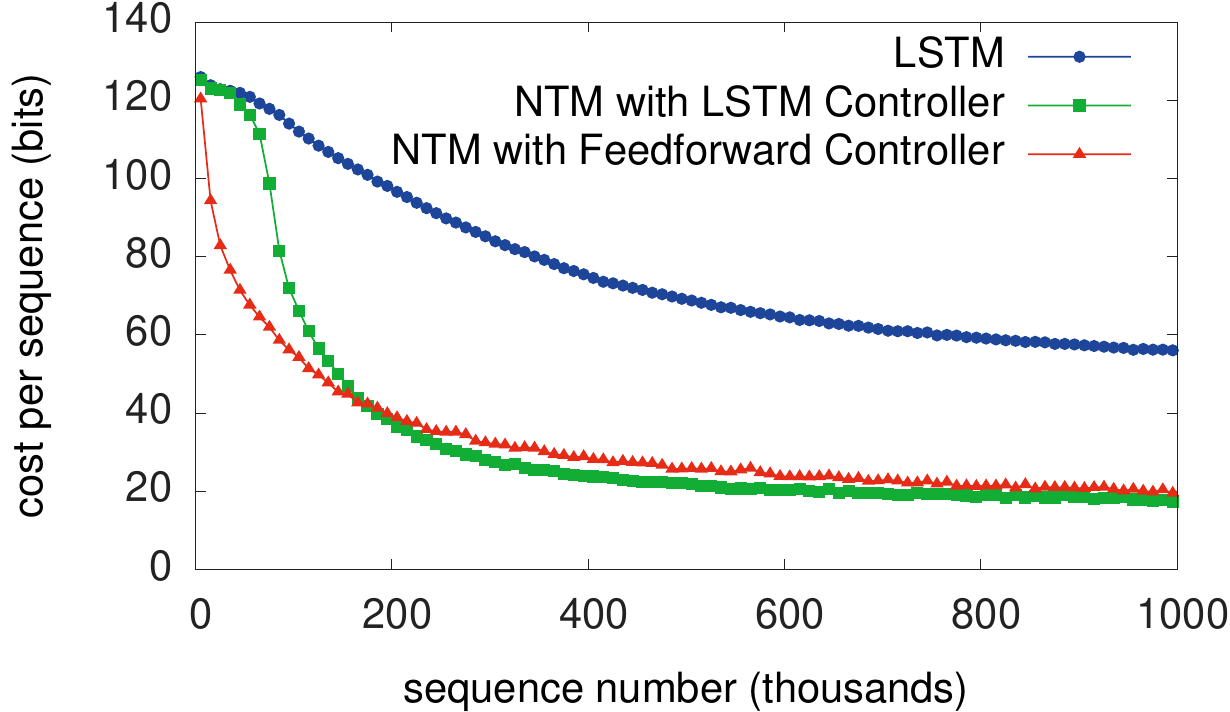}{sort_cost}{\costwidth}{Priority Sort Learning Curves}{}

\subsection{Experimental Details}\seclabel{details}
For all experiments, the \textit{RMSProp} algorithm was used for training in the form described in \citep{graves2013generating} with momentum of 0.9. \tref{details_ntm_ff,tab:details_ntm_lstm,tab:details_lstm} give details about the network configurations and learning rates used in the experiments. All LSTM networks had three stacked hidden layers. Note that the number of LSTM parameters grows quadratically with the number of hidden units (due to the recurrent connections in the hidden layers). This contrasts with NTM, where the number of parameters does not increase with the number of memory locations. During the training backward pass, all gradient components are clipped elementwise to the range (-10, 10). 

\begin{table}
\begin{small}
\centering 
\begin{tabular}{l c c c c r} 
\spaceline
Task & \#Heads & Controller Size & Memory Size & Learning Rate & \#Parameters \\ 
\spaceline 
Copy & 1 & 100 & 128 $\times$ 20 & $10^{-4}$ & $17,162$ \\ 
Repeat Copy & 1 & 100 & 128 $\times$ 20 & $10^{-4}$ & $16,712$ \\
Associative & 4 & 256 & 128 $\times$ 20 & $10^{-4}$ & $146,845$ \\
N-Grams & 1 & 100 & 128 $\times$ 20 & $3 \times 10^{-5}$ & $14,656$ \\
Priority Sort & 8 & 512 & 128 $\times$ 20  & $3 \times 10^{-5}$ & $508,305$ \\  
\spaceline 
\end{tabular}
\caption{\small{\textbf{NTM with Feedforward Controller Experimental Settings}}} 
\tlabel{details_ntm_ff} 
\end{small}
\end{table}

\begin{table}
\begin{small}
\centering 
\begin{tabular}{l c c c c r} 
\spaceline 
Task & \#Heads & Controller Size & Memory Size & Learning Rate & \#Parameters \\ 
\spaceline 
Copy & 1 & 100 & 128 $\times$ 20 & $10^{-4}$ & $67,561$ \\ 
Repeat Copy & 1 & 100 & 128 $\times$ 20 & $10^{-4}$ & $66,111$ \\
Associative & 1 & 100 & 128 $\times$ 20 & $10^{-4}$ & $70,330$ \\
N-Grams & 1 & 100 & 128 $\times$ 20 & $3 \times 10^{-5}$ & $61,749$ \\
Priority Sort & 5 & $2 \times 100$ & 128 $\times$ 20  & $3 \times 10^{-5}$ & $269,038$ \\ 
\spaceline 
\end{tabular}
\caption{\small{\textbf{NTM with LSTM Controller Experimental Settings}}} 
\tlabel{details_ntm_lstm} 
\end{small}
\end{table}

\begin{table}
\begin{small}
\centering 
\begin{tabular}{l c c r} 
\spaceline
Task & Network Size & Learning Rate & \#Parameters \\  
\spaceline 
Copy & $3 \times 256$ & $3 \times 10^{-5}$ & $1,352,969$ \\ 
Repeat Copy & $3 \times 512$ & $3 \times 10^{-5}$ & $5,312,007$ \\
Associative & $3 \times 256$ & $10^{-4}$ & $1,344,518$ \\
N-Grams & $3 \times 128$ & $10^{-4}$ & $331,905$ \\
Priority Sort & $3 \times 128$ & $3 \times 10^{-5}$ & $384,424$ \\  
\spaceline 
\end{tabular}
\caption{\small{\textbf{LSTM Network Experimental Settings}}} 
\tlabel{details_lstm} 
\end{small}
\end{table}

\section{Conclusion}
We have introduced the Neural Turing Machine, a neural network architecture that takes inspiration from both models of biological working memory and the design of digital computers.
Like conventional neural networks, the architecture is differentiable end-to-end and can be trained with gradient descent.
Our experiments demonstrate that it is capable of learning simple algorithms from example data and of using these algorithms to generalise well outside its training regime.

%
%


\section{Acknowledgments} 
Many have offered thoughtful insights, but we would especially like to thank Daan Wierstra, Peter Dayan, Ilya Sutskever, Charles Blundell, Joel Veness, Koray Kavukcuoglu, Dharshan Kumaran, Georg Ostrovski, Chris Summerfield, Jeff Dean, Geoffrey Hinton, and Demis Hassabis.

\clearpage
\newpage

\bibliographystyle{apalike}
\bibliography{ntm_arxiv} 
                                                        
\end{document}